\documentclass[10pt,twocolumn,letterpaper]{article}
\usepackage{cvpr}
\usepackage{times}
\usepackage{epsfig}
\usepackage{graphicx}
\usepackage{amsmath}
\usepackage{amssymb}
\usepackage{threeparttable}
\usepackage{multirow}
\usepackage{float}
\usepackage{multicol}
\usepackage{graphicx}
\usepackage{array,multirow}
\usepackage{setspace}
\usepackage{eucal}
\usepackage[bookmarks=false]{hyperref}
\usepackage{xcolor}
\usepackage{caption}
\hypersetup{
	colorlinks,
	linkcolor={red!70!black},
	citecolor={blue!50!black},
	urlcolor={blue!80!black}
}

\cvprfinalcopy 

\def\cvprPaperID{} 

\ifcvprfinal\fi
\begin{document}
	
	\begin{multicols}{2}
		\title{Joint 2D-3D-Semantic Data for Indoor Scene Understanding}
		
		\author{Iro Armeni$^1$\thanks{Both authors contributed equally.} \;\; Alexander Sax$^1$\footnotemark[1] \;\; Amir R. Zamir$^{1,2}$ \;\; Silvio Savarese$^1$\\ \vspace{2mm}
			$^1$ Stanford University \;\; $^2$ University of California, Berkeley \\ \vspace{2mm}
			\textcolor{blue}{\url{http://3Dsemantics.stanford.edu/}}\vspace{-5mm}
		}


	\end{multicols}

	\twocolumn[{%
		\renewcommand\twocolumn[1][]{#1}%
		\maketitle
		\begin{center}
			\centering
			\vspace{-4mm}
			\includegraphics[width=\textwidth]{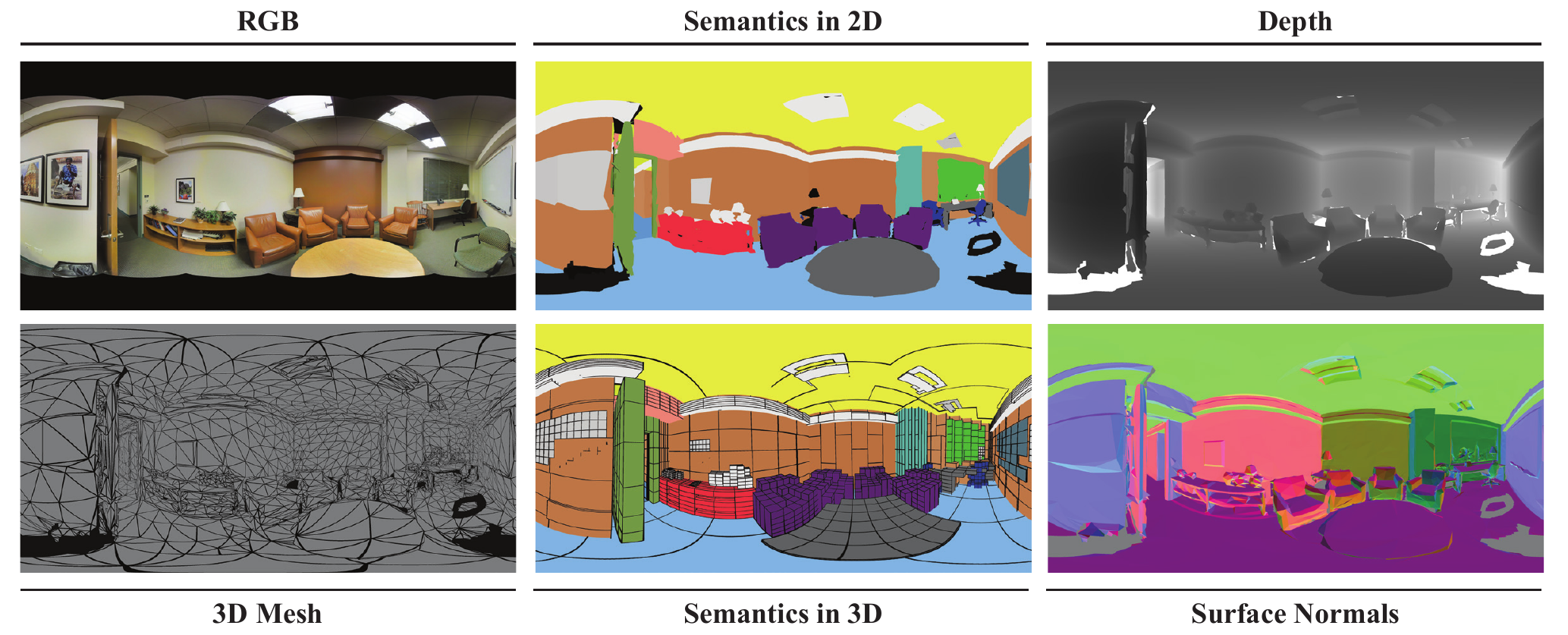}
		\end{center}%
		\vspace{-6mm}
		\captionof{figure}{\footnotesize{\textbf{ Joint 2D-3D-Semantic data.} We present a dataset that provides a variety of mutually registered modalities including: RGB images, depth, surface normals, global XYZ images as well as instance-level semantic annotations in 2D and 3D. The modalities are in 2D (\eg RGB images), 2.5D (\eg depth) and 3D (\eg meshes).}
		}%
		\label{fig:teaser}
		\vspace{5mm}}]

	\begin{abstract}
		We present a dataset of large-scale indoor spaces that provides a variety of mutually registered modalities from 2D, 2.5D and 3D domains, with instance-level semantic and geometric annotations. The dataset covers over 6,000~$m^2$ and contains over 70,000 RGB images, along with the corresponding depths, surface normals, semantic annotations, global XYZ images (all in forms of both regular and $360^{\circ}$~equirectangular images) as well as camera information. It also includes registered raw and semantically annotated 3D meshes and point clouds. The dataset enables development of joint and cross-modal learning models and potentially unsupervised approaches utilizing the regularities present in large-scale indoor spaces.
	\end{abstract}
	
	\footnote[0]{$^*$ Both authors contributed equally.}
	
	\begin{table*}
		\centering
		\caption{\footnotesize{Comparison of existing 2.5D and 3D Datasets.}}
		\vspace{0mm}
		\resizebox{\textwidth}{!}{
			\begin{tabular}{c|cccc|ccc|c}
				\textbf{Dataset}   & \textbf{Stanford Scenes}~\cite{fisher2012example} & \textbf{SceneNet}~\cite{Handa_2016_CVPR}& \textbf{SceneNet RGBD}~\cite{mccormac2016scenenet} & \textbf{SUNCG}~\cite{song2016semantic} & \textbf{NYUD2}~\cite{Silberman:ECCV12} & \textbf{SUN RGBD}~\cite{song2015sun} & \textbf{SceneNN}~\cite{hua2016scenenn} & \textbf{2D-3D-S (Ours)} \\
				Type of Data                        & Synthetic                & Synthetic             & Synthetic                   & Synthetic                & Real                      & Real               & Real                  & Real  \\ \hline
				RGB                                 & -                        & -                     & 5M                          & -                        & 1,449                     & 10,335             & -   & 70,496     \\
				Depth                               & \includegraphics[width=0.1in]{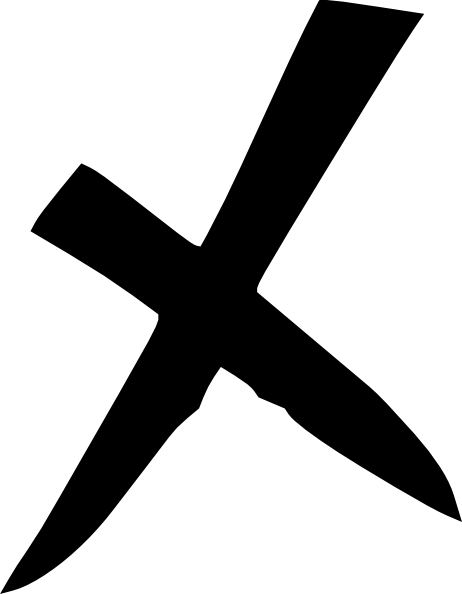} & \includegraphics[width=0.1in]{figures/x.png} & \includegraphics[width=0.1in]{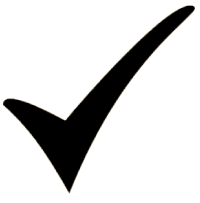} & 130,269 & \includegraphics[width=0.1in]{figures/Tick.png} &  \includegraphics[width=0.1in]{figures/Tick.png}  & \includegraphics[width=0.1in]{figures/Tick.png} & \includegraphics[width=0.1in]{figures/Tick.png}  \\
				Collection Method                & \includegraphics[width=0.1in]{figures/x.png} & \includegraphics[width=0.1in]{figures/x.png} & Rendered Video & Rendered Depth Images & Video &  Video  & Video & $360^{\circ}$ scan \\
				Surf. Normals                       & \includegraphics[width=0.1in]{figures/x.png} & \includegraphics[width=0.1in]{figures/x.png} & \includegraphics[width=0.1in]{figures/x.png}    & \includegraphics[width=0.1in]{figures/x.png} & \includegraphics[width=0.1in]{figures/Tick.png} &  \includegraphics[width=0.1in]{figures/Tick.png}  & \includegraphics[width=0.1in]{figures/Tick.png} & \includegraphics[width=0.1in]{figures/Tick.png}\\
				2D Semantics                        & \includegraphics[width=0.1in]{figures/x.png} & \includegraphics[width=0.1in]{figures/x.png} & \includegraphics[width=0.1in]{figures/Tick.png} & \includegraphics[width=0.1in]{figures/x.png} & \includegraphics[width=0.1in]{figures/Tick.png} &  \includegraphics[width=0.1in]{figures/Tick.png}  & \includegraphics[width=0.1in]{figures/Tick.png} & \includegraphics[width=0.1in]{figures/Tick.png}\\
				Resolution              & -                                            & -                                            & $320\times240$                                         & -                                            & $640\times480$                            &  $640\times480$                                          & $640\times480$                                         & $1080\times1080$\\
				\hline
				3D Point Cloud (PC)                 & \includegraphics[width=0.1in]{figures/x.png} & \includegraphics[width=0.1in]{figures/x.png} & \includegraphics[width=0.1in]{figures/x.png}    & \includegraphics[width=0.1in]{figures/x.png} & \includegraphics[width=0.1in]{figures/Tick.png} & \includegraphics[width=0.1in]{figures/Tick.png} & \includegraphics[width=0.1in]{figures/x.png} & \includegraphics[width=0.1in]{figures/Tick.png}  \\
				3D Mesh /CAD                        & \includegraphics[width=0.1in]{figures/Tick.png} & \includegraphics[width=0.1in]{figures/Tick.png} & \includegraphics[width=0.1in]{figures/x.png} & \includegraphics[width=0.1in]{figures/Tick.png} & \includegraphics[width=0.1in]{figures/x.png} & \includegraphics[width=0.1in]{figures/x.png} & \includegraphics[width=0.1in]{figures/Tick.png} & \includegraphics[width=0.1in]{figures/Tick.png}  \\
				3D Semantic Mesh/ CAD               & \includegraphics[width=0.1in]{figures/Tick.png} & \includegraphics[width=0.1in]{figures/Tick.png} & \includegraphics[width=0.1in]{figures/x.png} & \includegraphics[width=0.1in]{figures/Tick.png} & \includegraphics[width=0.1in]{figures/x.png} & \includegraphics[width=0.1in]{figures/x.png} & \includegraphics[width=0.1in]{figures/Tick.png} & \includegraphics[width=0.1in]{figures/Tick.png}  \\
				\hline
				\# Object Class                     & -   & -   & 255  & 84       & 894 & 800 & -    & 13\\ 
				\# Scene Categories                 & -   & 5   & 5    & 24       & 26  & 47  & -    & 11 \\ 
				\# Scene Layouts                    & 130 & 57  & 57*  & 45,622  & 464 & -   & 100  & 270 \\
				\hline
				\hline
				\multicolumn{9}{c}{\includegraphics[width=0.1in]{figures/x.png}: not included, \includegraphics[width=0.1in]{figures/Tick.png}: included, \textbf{-}: information not available, *:16,895 configurations}
			\end{tabular}
		}
		\label{tab:comparison}
	\end{table*}
	\vspace{-8mm}

	\section{Introduction} \label{sec:intro}
	
	There is an important advantage in analyzing 3D data,  especially ones that originate from comprehensive large-scale scans: the entire geometry of an object and its surrounding context are available at once. This can provide strong cues for semantics, layout, occlusion  handling, shape completion, amodal detection, etc. This rich geometric information is complementary to the RGB domain, which offers dense appearance features. Hence, there is a great potential in developing models that perform joint or cross-modal learning to enhance the performance. Although several RGB-D~\cite{Silberman:ECCV12, song2015sun, hua2016scenenn, mccormac2016scenenet} and a few 3D datasets~\cite{fisher2012example, Handa_2016_CVPR, song2016semantic, armeni_cvpr16} have been developed to date, the majority of them are limited in the scale, diversity, and/or the number of modalities they provide. 
	
	Over the past few years, advances in the field of 3D imaging have led to manufacturing inexpensive sensors and mainstreaming their use in consumer products (\eg Kinect~\cite{Kinect}, Structure Sensor~\cite{StructureSensor}, RealSense~\cite{RealSense}, etc). The Computer Vision community has been affected by this change and is experiencing a lot of development in data-driven 2.5D and 3D vision~\cite{Silberman:ECCV12, ren2012rgb, bo2011object, koppula2011semantic, armeni_cvpr16}. However, the 3D sensing field has recently undergone a follow-up shift with the availability of mature technology for scanning \textit{large-scale} spaces, \eg an entire building. Such systems can reliably form the 3D point cloud of thousands of square meters with the number of points often exceeding hundreds of millions. This demands developing methods capable of coping with this scale, and ideally, exploiting the unique characteristics of such data.
	
	To enable parsing the aforementioned goals, we present a 2D-3D semantic dataset that can be used for a plethora of tasks, such as scene understanding, depth estimation, surface normals estimation, object detection, segmentation, amodal detection, and scene reconstruction. Also, the 3D mesh models and equirectangular projections can be used to generate a virtually unlimited number of images, something that is currently possible only in synthetic 3D datasets. The dataset along with $360^{\circ}$ visualizations is available for download at \textcolor{blue}{\href{http://3Dsemantics.stanford.edu/}{http://3Dsemantics.stanford.edu/}}.

	\section{Related Datasets}
	
	There exist several RGB-D datasets in the literature related to scene understanding; NYU Depth v2\cite{Silberman:ECCV12}, SUN RGBD\cite{song2015sun} and recently SceneNN~\cite{hua2016scenenn} are among the prominent ones. The latter provides a larger number of images than the rest, though the increased number originates from densely annotated frames of videos from a smaller number of scenes. Due to the complexity of collecting and densely annotating such data, many synthetic datasets have appeared lately (RGBD:~\cite{mccormac2016scenenet}, 3D:~\cite{fisher2012example, Handa_2016_CVPR, song2016semantic}). Their advantage is that by employing large-scale object libraries (\eg ShapeNet~\cite{chang2015shapenet}) one could generate a virtually infinite number of images along with the corresponding semantics. 
	
	Existing real datasets are relatively limited to one particular task and the 2.5D domain. Recently, \cite{hua2016scenenn} offered watertight mesh models of the reconstructed scenes and \cite{armeni_cvpr16} presented a 3D point cloud dataset of large-scale indoor spaces. Although such data can be used for more than semantic detection/segmentation, it is not suitable for tasks across different data dimensionalities and modalities.
	
	The proposed 2D-3D dataset includes RGB, depth, equirectangular and global XYZ OpenEXR images, as well as 3D meshes and point clouds of the same indoor spaces (Table~\ref{tab:comparison}). The different modalities can be used independently or jointly to develop learning models that seamlessly transcend across domains. Also, using the provided equirectangular images, camera parameters, and 3D mesh models, it is possible to generate additional data tailored to specific tasks. In comparison to existing real-world datasets, it offers a greater number of images as well as additional modalities, such that of equirectangular images or 3D meshes. It also provides consistent annotations across all modalities and dimensions. However, it is currently limited in the number of object and scene categories.

	\begin{figure*}
		\centering
		\setcounter{figure}{1}
		\centerline{\includegraphics[width=1\textwidth]{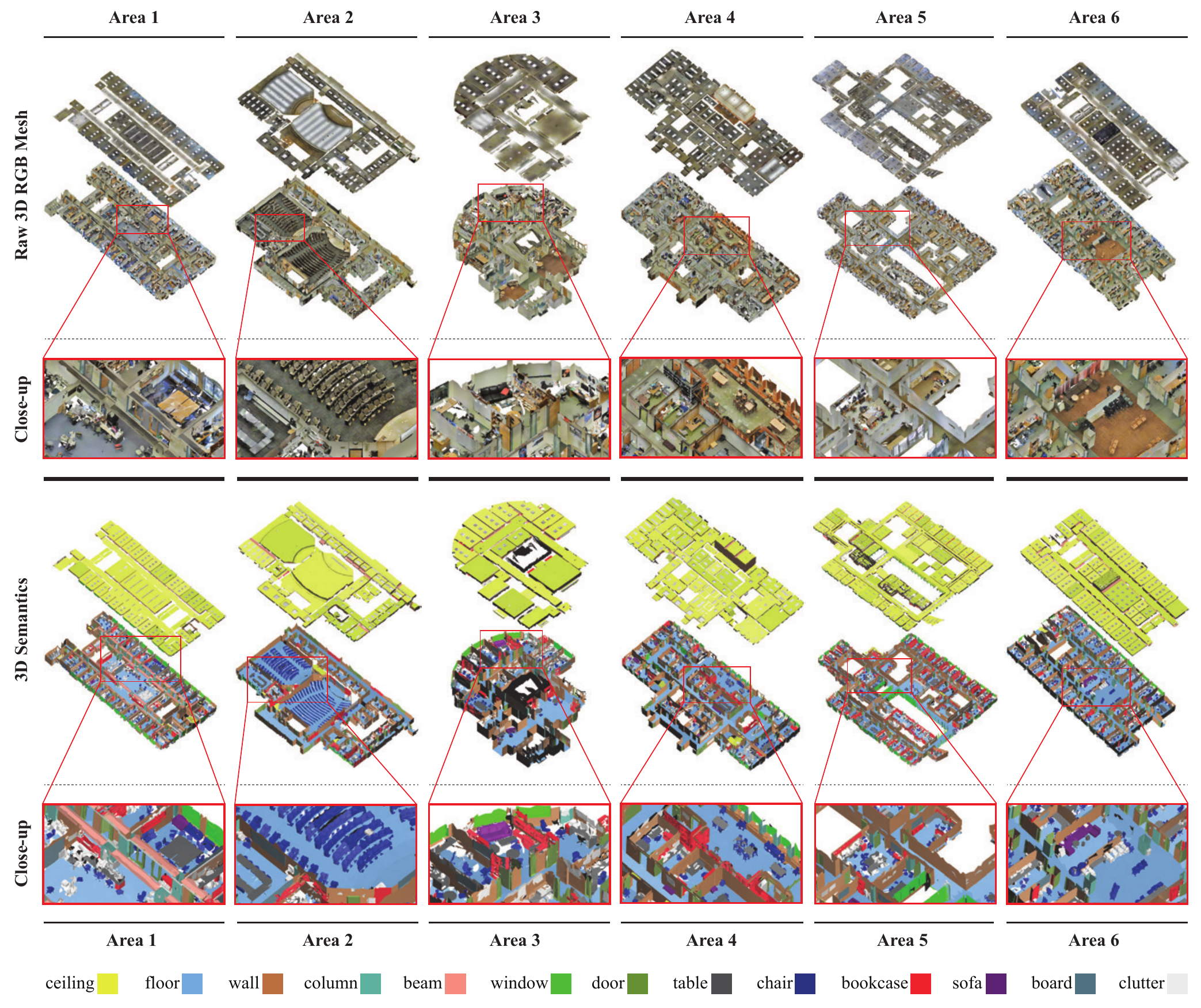}}
		\caption{\footnotesize{\textbf{3D Modalities.} The dataset includes both the textured and semantic 3D mesh models of all areas as well as their point clouds.}}
		\label{fig:3Dsemantics}
	\end{figure*}

	\section{Dataset Overview}
	
\begin{figure*}
		\centering
		\centerline{\includegraphics[width=1\textwidth]{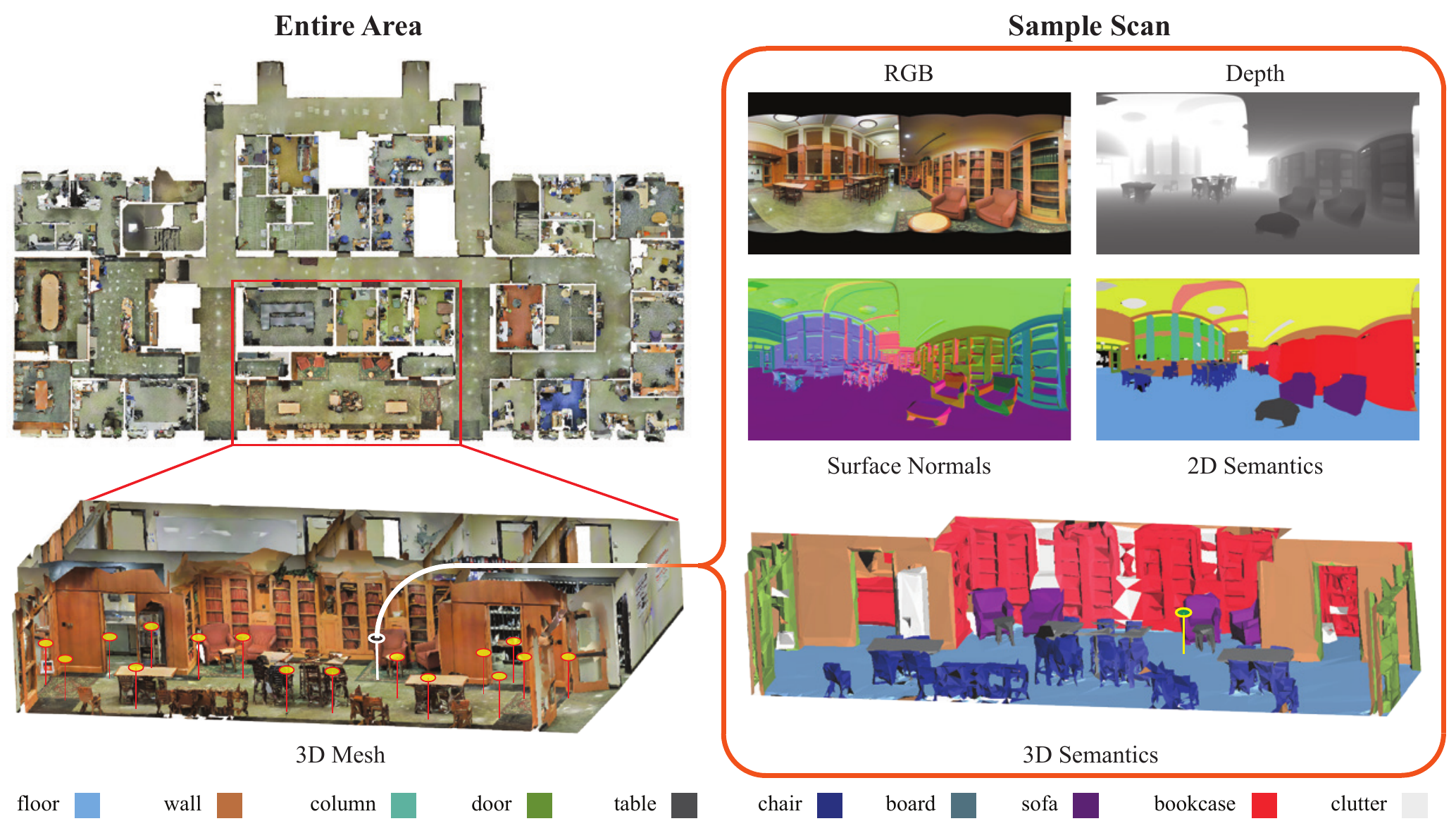}}
		\caption{\footnotesize{\textbf{Data Processing.} The RGB output of the scanner is registered on the 3D modalities (each yellow marker represents one scan). We then process the RGB and 3D data to make depth, surface normals, and 2D semantic (projected from 3D semantics) images for each scan.  The processed equirectangular images are sampled to make new regular images (shown in Figure~\ref{fig:sampling}).}}
		\label{fig:panorama}
	\end{figure*}

	The dataset is collected in 6 large-scale indoor areas that originate from 3 different buildings of mainly educational and office use. For each area, all modalities are registered in the same reference system, yielding pixel to pixel correspondences among them. In a nutshell, the presented dataset contains a total of 70,496 regular RGB and 1,413 equirectangular RGB images, along with their corresponding depths, surface normals, semantic annotations, global XYZ OpenEXR format and camera metadata. In addition, we provide whole building 3D reconstructions as textured meshes, as well as the corresponding 3D semantic meshes. We also include the colored 3D point cloud data of these areas with the total number of 695,878,620 points, that has been previously presented in the Stanford large-scale 3D Indoor Spaces Dataset (S3DIS~\cite{armeni_cvpr16}).  The annotations are instance-level, and consistent across all modalities and correspond to 13 object classes. We refer the readers to Tables \ref{tab:sceneClassStats} and \ref{tab:objectStats} for statistics on the scene and object categories. Figure~\ref{fig:panorama} shows an example of the equirectangular images for one scan. 
	
	
	\section{Collection and Processing}
	
	We collected the data using the Matterport Camera~\cite{matterport}, which combines 3 structured-light sensors to capture 18 RGB and depth images during a $360^{\circ}$ rotation at each scan location. The output is the reconstructed 3D textured meshes of the scanned area, the raw RGB-D images, and camera metadata. We used this data as a basis to generate additional RGB-D data and make point clouds by sampling the meshes. We semantically annotated the data directly on the 3D point cloud, rather than images, and then projected the per point labels on the 3D mesh and the image domains. 
	
	The rest of the section elaborates on each modality.
	
	\subsection{3D modalities}
	The dataset contains two main 3D modalities (3D point cloud data and 3D mesh model) and their semantic counterparts for each of the 6 areas. Statistics related to this modality are offered in Table \ref{tab:3DmodalStats}.

	\textbf{3D Point Cloud and Mesh:} 
	As mentioned above, we receive the reconstructed 3D textured Mesh model for each scanned area from the Matterport Camera. Each model contains an average of 200k triangulated faces and a material mapping to texture images providing a realistic reconstruction of the scanned space. We generate the colored 3D point clouds by densely and uniformly sampling points on the mesh surface and assigning the corresponding color. 
	

	\begin{table}
		\centering
		\caption{\footnotesize{Statistics of 3D Data}}
		\vspace{-3mm}
		\resizebox{0.90\columnwidth}{!}{
			\begin{tabular}{c|cc}
				\small{Area}&
				 \small{~Number of 3D Points} & \small{~Number of Mesh Faces} \\
				\hline
				\small{1} & \small{43,956,907}  & \small{158,500}\\
				\small{2} & \small{470,023,210} & \small{361,830}\\
				\small{3} & \small{18,662,173}  & \small{147,420}\\
				\small{4} & \small{43,278,148}  & \small{201,735}\\
				\small{5} & \small{78,649,818}  & \small{198,220}\\
				\small{6} & \small{41,308,364}  & \small{198,590}\\
				\hline
				\textbf{Total} & \textbf{\small{695,878,620}} & \textbf{\small{1,266,295}}\\
			\end{tabular}
		}
		\label{tab:3DmodalStats}
	\end{table}
		
	\textbf{3D Semantics (Labeled Mesh and Voxels):}
	We semantically annotate the data on the \emph{3D point cloud} and assign one of the following 13 object classes on a per-point basis: \textit{ceiling, floor, wall, beam, column, window, door, table, chair, sofa, bookcase, board} and \textit{clutter} for all other elements. Performing annotations in 3D, rather than 2D, provides 3D object models and enables performing occlusion and amodal analysis, yet the semantics can be projected onto any number of images to provide ground truth annotations in 2D as well. Each object instance in the dataset has a unique identifier. We also annotate the point cloud data into rooms and assign one of the following 11 scene labels to each: \textit{office, conference room, hallway, auditorium, open space, lobby, lounge, pantry, copy room, storage} and \textit{WC}. Again, each instance in the point cloud receives a unique index. Given these annotations, we calculate the tightest axis-aligned object bounding box of each instance and further voxelize it into a $6\times6\times6$ grid with binary occupancy. This information provides a better understanding of the underlying geometry and can be leveraged, for example, in 3D object detection or classification.  
	
	We then project the object and scene semantics on the mesh model's faces and generate 3D semantic meshes that preserves the same class structure and instance index. We used a voting scheme to transfer these annotations to the mesh. Each annotated point casts a vote for the face that is nearest to it, then votes are tallied and each face is annotated with the mode class. Faces which garner no votes belong to non-annotated parts of the dataset and are labeled as the default \textit{\textless UNK\textgreater\_0\_\textless UNK\textgreater\_0\_0} class (null). Our 3D models are labeled with the class of the object and the specific instance. These instances are globally unique among all models and are indexed in \textit{$semantic\_labels.json$}. Each object is stored in this file as \textit{$class\_instanceNum\_roomType\_roomNum\_areaNum$}. Note that instances, rooms and areas are 1-indexed so that the singleton \textit{\textless UNK\textgreater} class is unique in that it has $instanceNum=0$, $roomNum=0$  and $areaNum=0$. Figure~\ref{fig:3Dsemantics} shows the raw and semantically annotated 3D mesh models for all 6 areas.

	\begin{figure*}
		\centering
		\centerline{\includegraphics[width=1\textwidth]{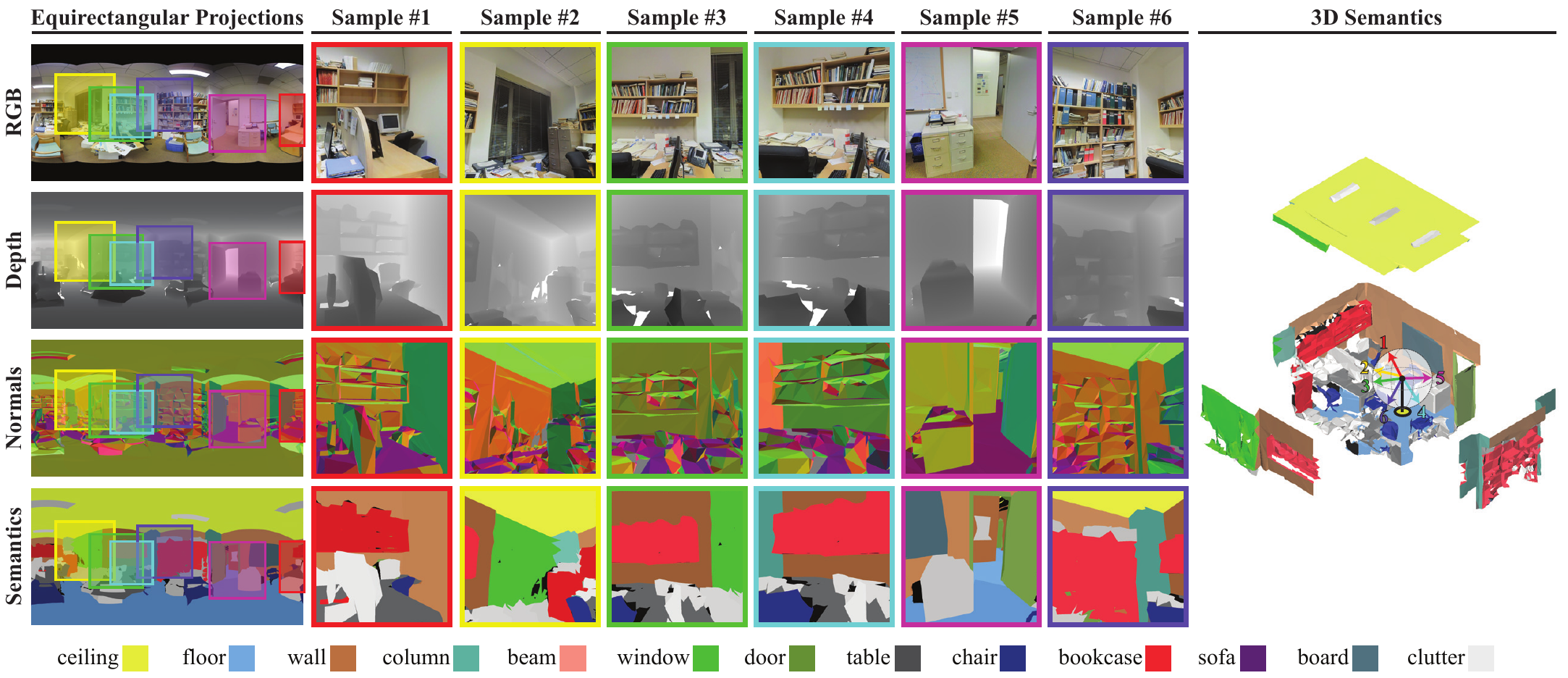}}
		\caption{\footnotesize{\textbf{Sampling images from the equirectangular projection.} We use the equirectangular projections to sample 72 images per scan location, all with consistent depth, surface normal, and semantic information. The sampling distributions are provided in Figure~\ref{fig:camera}.}}
		\label{fig:sampling}
	\end{figure*}
	
	\begin{figure}
		\centering
		\centerline{\includegraphics[width=1\columnwidth]{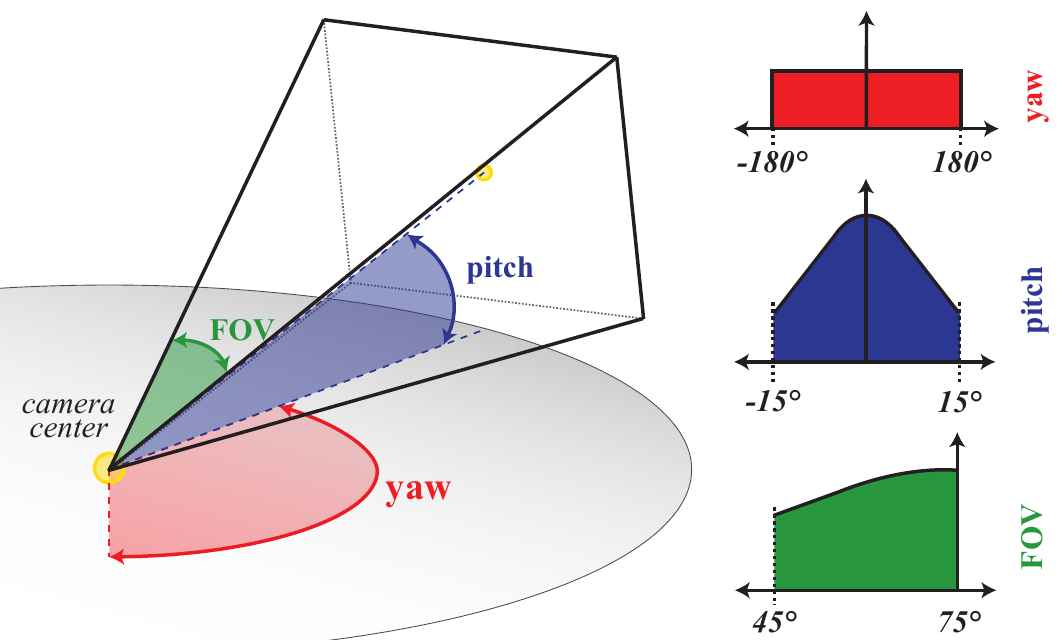}}
		\caption{\footnotesize{\textbf{Sampling distributions.} We sample camera parameters from the above distributions of yaw, pitch and Field of View (FOV).}}
		\label{fig:camera}
	\end{figure}

	\subsection{2D modalities}
	The dataset contains densely sampled RGB images per scan location. These images were sampled from equirectangular images that were generated per scan location and modality using the raw data captured by the scanner (also part of the dataset). Statistics of the 2D modalities are offered in Table \ref{tab:2DmodalStats}. All images in the dataset are stored in full high-definition at $1080\times1080$ resolution.

	\textbf{RGB Images}: 
	We use the provided raw RGB data to form a cubemap per scan location and sample new images in this space. We randomly sample the camera's euler angles as follows (Figure~\ref{fig:camera}): (a) yaw: uniform in $[-180^{\circ},180^{\circ}]$, (b) pitch: Gaussian with zero mean and $15^{\circ}$ standard deviation, and (c) roll: always zero. Field of View (FOV) angles are sampled from a half Gaussian distribution with $75^{\circ}$ mean and $-30^{\circ}$ standard deviation.

	To avoid having images that are uninteresting in terms of semantic content (\eg only a plain wall), we generate $3\times24$ images per scan location and preserve approximately 70\% of them by sampling based on semantic content entropy. We perform the entropy sampling as follows: we compute Shannon's entropy for each image on the pixel distribution of semantic classes therein (i.e. a 13 bin distribution). We then discard the bottom $\sim$15\% (as such cases correspond to images with very small semantic value, \eg a close up of a wall) and preserve the top $\sim$60\% (the semantically diverse images). Out of the rest of the images, we preserve about 50\% of them by sampling the entropy values on a half Gaussian with mean and standard deviation of 1 and -$\frac{1}{2}$, respectively.

	Using this approach, rather than simply discarding the low-entropy images, preserves the dataset's diversity by not completely removing low-entropy scenes. We rendered all images via Blender 2.78. Figure~\ref{fig:sampling} shows examples of sampled images per modality on an equirectangular image.

	\textbf{Meta-data and Camera Parameters per Image}:
	For each generated image we provide the camera pose in the `pose folder'.

	\begin{table*}
		\centering
		\caption{\footnotesize{Object Class Statistics}}
		\vspace{-3mm}
		\resizebox{0.85\textwidth}{!}{
			\begin{tabular}{c|ccccccc|ccccc|c}
				\multirow{3}{*}{\textbf{Area}} & \multicolumn{7}{c}{\textbf{Structural Elements}} & \multicolumn{5}{c}{\textbf{Movable Elements}} & \multirow{2}{*}{\textbf{Total}} \\
				& ceiling & floor & wall & beam & column & window & door & chair & table & bookcase & sofa & board & \\
				\hline
				1 & 56 & 45 & 235 & 62 & 58 & 30 & 87  & 156 & 70  & 91  & 7  & 28 & \textbf{925} \\
				2 & 82 & 51 & 284 & 12 & 20 & 9  & 94  & 546 & 47  & 49  & 7  & 18 & \textbf{1,219} \\
				3 & 38 & 24 & 160 & 14 & 13 & 9  & 38  & 68  & 31  & 42  & 10 & 13 & \textbf{460} \\
				4 & 74 & 51 & 281 & 4  & 39 & 41 & 108 & 160 & 80  & 99  & 15 & 11 & \textbf{963} \\
				5 & 77 & 69 & 344 & 4  & 75 & 53 & 128 & 259 & 155 & 218 & 12 & 43 & \textbf{1,437} \\
				6 & 64 & 50 & 248 & 69 & 55 & 32 & 94  & 180 & 78  & 91  & 10 & 30 & \textbf{1,001} \\
				\hline
				\textbf{Total} & \textbf{391} & \textbf{290} & \textbf{1,552} & \textbf{165} & \textbf{260} & \textbf{174} & \textbf{549} & \textbf{1,369} & \textbf{461} & \textbf{590} & \textbf{61} & \textbf{143} & \textbf{6,005} \\
			\end{tabular}
		}
		\label{tab:objectStats}
	\end{table*}

	\textbf{Depth Images}:
	For each image, we provide the depth, which was computed from the 3D mesh instead of directly from the scanner. We rendered these images from the 3D mesh by saving out depth information from the z-buffer in Blender. The images are saved as 16-bit grayscale PNGs where one unit of change in pixel intensity (e.g. a value from 45 to 46) corresponds to a $\frac{1}{512}m$ change in depth. The maximum observable range is therefore about 128 meters ($2^{16}\cdot\frac{1}{512}$). All depths beyond this maximum distance assume the maximum value (65,535). Pixels that correspond to locations where there is no depth information also take this maximum distance. 

	\textbf{Surface Normal Images}:
	The surface normals were computed from a normals pass in Blender and are saved as 24-bit RBG PNGs. The surface normals in 3D corresponding to each pixel are computed from the 3D mesh instead of directly from the depth image. The normal vector is saved in the RGB color value where Red is the horizontal value (more red to the right), Green is vertical (more green downwards), and Blue is towards the camera. Each channel is 127.5-centered, so both values to the left and right (of the axis) are possible. For example, a surface normal pointing straight at the camera would be colored (128, 128, 255) since pixels must be integer-valued. 
	
	Missing values take (128,128,128) which is convenient in practice as it is not a unit normal and is clearly visually distinguishable from the surrounding values. The convention is that surface normals cannot point away from the camera.

	\begin{table}
		\centering
		\caption{\footnotesize{Statistics of Images}}
		\vspace{-2mm}
		\resizebox{0.75\columnwidth}{!}{
			\begin{tabular}{c|cc|c}
				\multirow{2}{*}{Area} & \multicolumn{2}{c}{\textbf{\# of Images per 2/2.5D Modality}} & \multirow{2}{*}{\textbf{Total}} \\
				& \small{Image Type \textbf{I}}  & \small{Image Type \textbf{E}} & \\
				\hline
				1 & 10,327 & 190 & 42,068\\
				2 & 15,714 & 299 & 64,052\\
				3 & 3,704  & 85  & 15,156\\
				4 & 13,268 & 258 & 54,104\\
				5 & 17,593 & 373 & 71,864\\
				6 & 9,890  & 208 & 40,392\\
				\hline
				\textbf{Total}& 70,496 & 1,413 &  287,636\\
				\hline
				\hline
				\multicolumn{4}{c}{\textbf{I}: Regular Images, \textbf{E}: Equirectangular Images\vspace{-0mm}}
			\end{tabular}}
			\label{tab:2DmodalStats}
			\vspace{-0mm}
		\end{table}

		\begin{table} 
			\centering
			\caption{\footnotesize{\textbf{Training and Testing Splits (3-fold cross-validation)}}}
			\vspace{-2mm}
			\resizebox{0.55\columnwidth}{!}{
				\begin{tabular}{c|cc}
					\multirow{2}{*}{Fold \#} & Training & Testing \\
					& (Area \#) & (Area \#) \\
					\hline
					1 & 1, 2, 3, 4, 6 & 5 \\
					2 & 1, 3, 5, 6    & 2, 4 \\
					3 & 2, 4, 5       & 1, 3, 6\\
				\end{tabular}
			}
			\label{tab:splits}
		\end{table}

		{\small
			\tabcolsep= 1mm
			\begin{table*}
				\centering
				\caption{\footnotesize{\textbf{Baseline 3D Object Detection Results (\cite{armeni_cvpr16})}. Class specific average precision (AP) using different features.}}
				\vspace{0mm}
				\resizebox{\textwidth}{!}{%
					\begin{tabular}{r|cccccccc|cccccc|c}
						& \multicolumn{8}{|c|}{\small{\textbf{Structural Elements}}} & \multicolumn{6}{|c|}{\small{\textbf{Movable Elements}}} & \small{\textbf{overall}}
						\\
						& \small{ceiling} & \small{floor} & \small{wall} & \small{beam} & \small{column} & \small{window} & \small{door} & \small{mean} & \small{table} & \small{chair} & \small{sofa} & \small{bookcase} & \small{board} & \small{mean} & \small{mean}
						\\
						Full model& \textbf{71.61} & \textbf{88.70} & \textbf{72.86} & 66.67 & \textbf{91.77} & \textbf{25.92} & \textbf{54.11} & \textbf{67.38} & 46.02 & \textbf{16.15} & \textbf{6.78} & 54.71 & 3.91 & \textbf{25.51} &  \textbf{49.93} \\
						No global & 48.93 & 83.76 & 65.25 & 62.86 & 83.15 & 22.55 & 41.08 & 57.27 & 37.57 & 11.80 & 4.57 & 45.49 & 3.06 & 20.35 & 41.87 \\
						No local & 50.74 & 80.48 & 65.59 & \textbf{68.53} & 85.08 & 21.17 & 45.39 & 58.73 & 39.87 & 11.43 & 4.91 & \textbf{57.76} & 3.73 & 23.78 & 44.19 \\ 
						No color & 48.05 & 80.95 & 67.78 & 68.02 & 87.41 & 25.32 & 44.31 & 59.73 & \textbf{50.56} & 11.83 & 6.32 & 52.33 & \textbf{4.76} & 25.30 & 45.41 \\
					\end{tabular}
				}
				\label{tab:CVPR16quantitative}
			\end{table*}
		}

		\textbf{Semantically Labeled Images}:
		We project the 3D semantics from the mesh model onto the 2D image. Due to certain geometric artifacts present at the mesh model mainly because of the level of detail in the reconstruction, the 2D annotations occasionally present small local misalignment to the underlying pixels, especially for points that have a short distance to the camera. This issue can be easily addressed by fusing image content with the projected annotations using graphical models.

		The semantically labeled images are saved as 24-bit RGB PNGs, but each pixel's color value can be directly interpreted as an index into the list. For example, \textit{$board\_3\_hallway\_4\_6$} is at index $257$ in \textit{$semantics\_labels.json$}. Since $257$ equals $\#000101$ in hex, $\#000101$ is the color of the chair in the image. For semantic images, pixel values that correspond to holes in the mesh contain the value (13, 13, 13).
		
		\textbf{3D Coordinate Encoded Images}:
		The pixels in these images encode the X, Y, Z location of the point in the world coordinate system. This information can be used for conveniently relating the content of the RGB images, \eg forming correspondences. The images are stored in the OpenEXR format with each channel contiaining 16-bit floating point numbers. Utility functions, including for reading EXRs, are provided in the GitHub repo accessible on the website.

		\subsection{Naming Convention}
        \includegraphics[width=0.89\columnwidth]{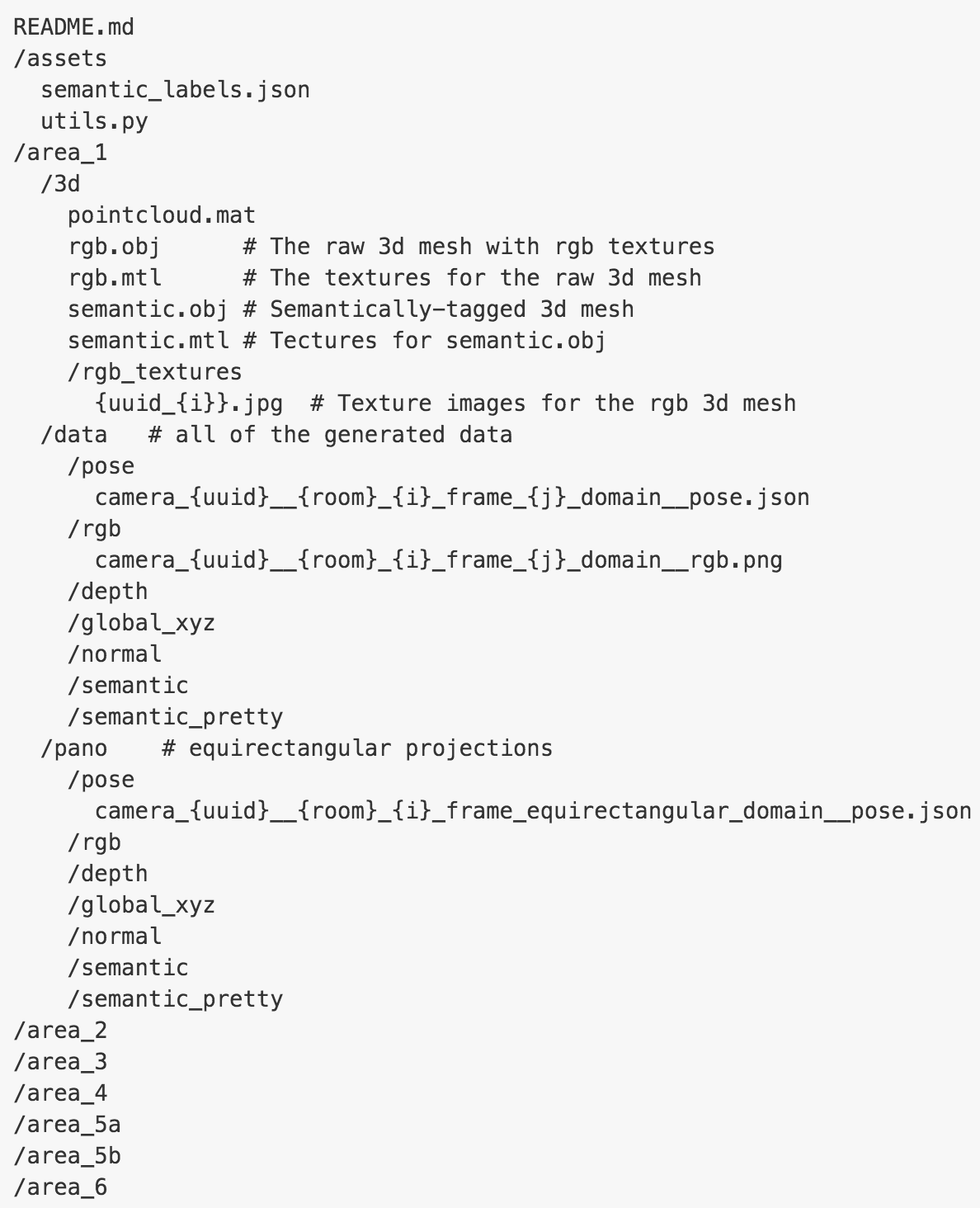}
		The filenames of images in the dataset are globally unique as no two files share a camera uuid, frame number, and domain. The room type is included for convenient filtering.

		\begin{figure*}
			\centering
			\centerline{\includegraphics[width=1\textwidth]{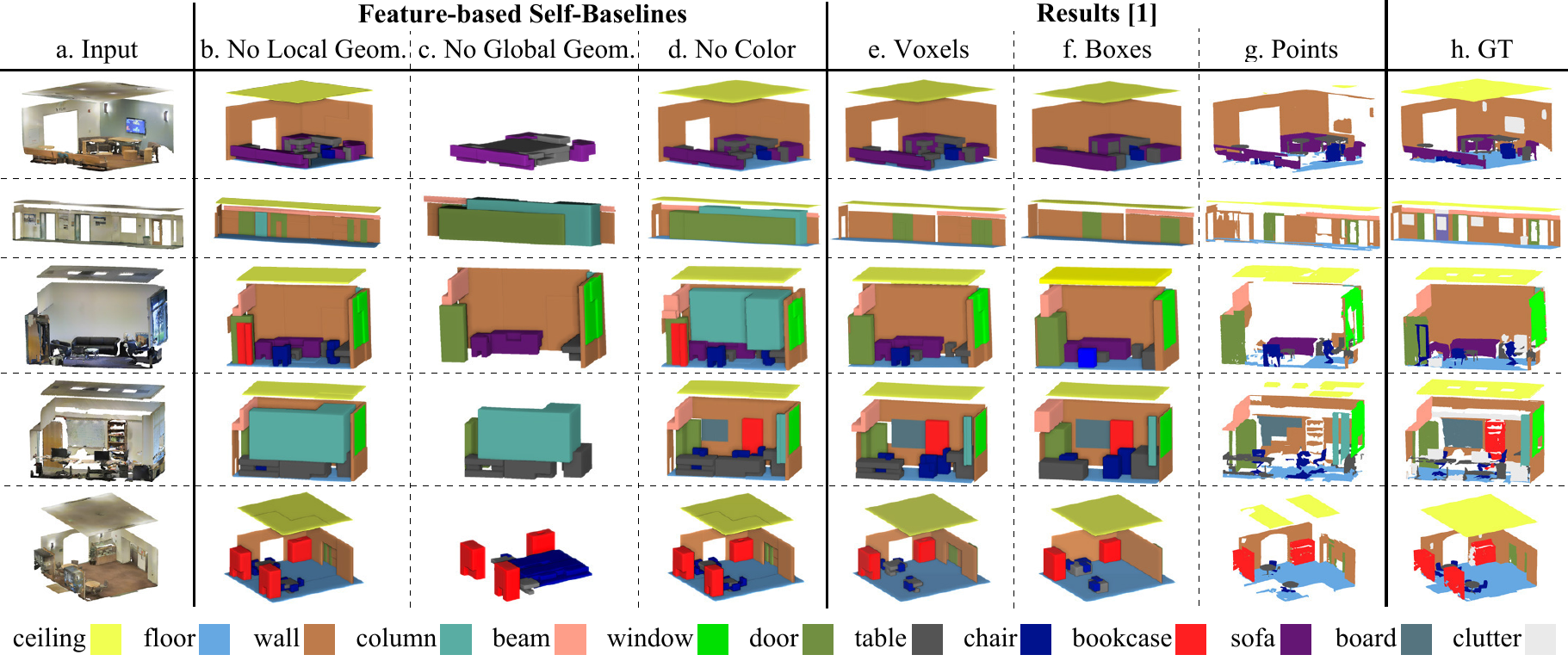}}
			\caption{\footnotesize{\textbf{Qualitative Baseline 3D Object Detection Results (\cite{armeni_cvpr16}).}}}
			\label{fig:CVPR16qualitative}
		\end{figure*}

		\begin{table*}
			\centering
			\caption{\footnotesize{Scene and Space Statistics}}
			\vspace{0mm}
			\resizebox{0.90\textwidth}{!}{
				\begin{tabular}{c|c|ccccccccccc|c}
					\multirow{3}{*}{\textbf{Area}} & \textbf{Sq.} & \multicolumn{11}{c}{\textbf{Number of Instances Per Scene Category}} & \multirow{3}{*}{\textbf{Total}} \\
					& \multirow{2}{*}{\textbf{meters}} & \multirow{2}{*}{Office} & Conference & \multirow{2}{*}{Auditorium} & \multirow{2}{*}{Lobby} & \multirow{2}{*}{Lounge} & \multirow{2}{*}{Hallway} & Copy & \multirow{2}{*}{Pantry} & Open & \multirow{2}{*}{Storage} & \multirow{2}{*}{WC} & \\
					&       &    & Room &  &  &  &  & Room & & Space & & & \\ \hline
					1 & 965   & 31 & 2 & - & - & - & 8  & 1 & 1 & - & - & 1 & \textbf{45}\\
					2 & 1,100 & 14 & 1 & 2 & - & - & 12 & - & - & - & 9 & 2 & \textbf{39}\\
					3 & 450   & 10 & 1 & - & - & 2 & 6  & - & - & - & 2 & 2 & \textbf{24}\\
					4 & 870   & 22 & 3 & - & 2 & - & 14 & - & - & - & 4 & 2 & \textbf{49}\\
					5 & 1,700 & 42 & 3 & - & 1 & - & 15 & - & 1 & - & 4 & 2 & \textbf{55}\\
					6 & 935   & 37 & 1 & - & - & 1 & 6  & 1 & 1 & 1 & - & - & \textbf{53}\\ \hline
					\textbf{Total} & \textbf{6,020} & \textbf{156} & \textbf{11} & \textbf{2} & \textbf{3} & \textbf{3} & \textbf{61} & \textbf{2} & \textbf{3} & \textbf{1} & \textbf{19} & \textbf{9} & \textbf{270}
				\end{tabular}
			}
			\label{tab:sceneClassStats}
		\end{table*}

		\section{Sample Data}
		
\begin{figure*}[ht]
			\centering
			\centerline{\includegraphics[width=\textwidth]{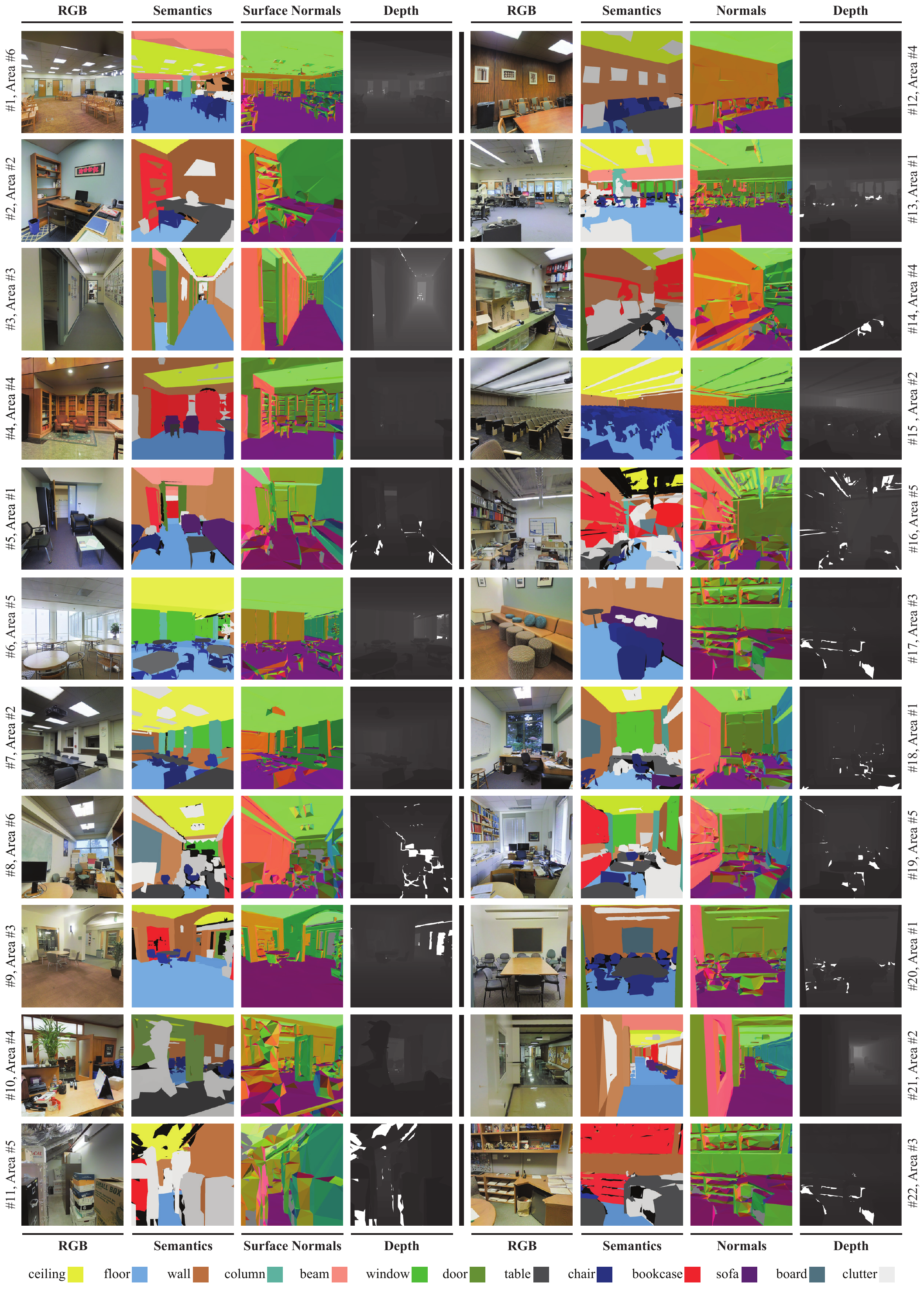}}
			\caption{\footnotesize{\textbf{Examples Images of 2D and 2.5D Modalities.} RGB, Semantic, Surface Normals and Depth images.}}
			\label{fig:examples}
		\end{figure*}

		Figure \ref{fig:examples} provides a representative sample of the generated data showing the diversity of the indoor scenes, in terms of scene category, appearance, intra-class variation, density of objects and amount of clutter. This also shows the varying degree of difficulty of the data, which consists of both easy and hard examples.

		\section{Train and Test splits}
		Certain areas in the dataset represent parts of buildings with similarities in their appearance and architectural features, thus we define standard training and testing splits so that no areas from similarly looking buildings appear in both. We split the 6 areas in the dataset as per Table~\ref{tab:splits} and follow a 3-fold cross-validation scheme.

		\section{Baseline Results}
		As a baseline, we provide results on the task of 3D object detection, performed on the 3D point clouds from this paper~\cite{armeni_cvpr16}. The method follows a hierarchical approach to semantic parsing of large-scale data: first, we parse the raw data into semantically meaningful spaces (\eg rooms, etc) and align them into a canonical reference coordinate system. Second, we parse each of these spaces into comprising elements that belong to one of the 12 available classes. We implement the first step with an unsupervised approach, and the second by training one-vs-all SVMs for each object class. We also employ a CRF for contextual consistency. Our experimental setup follows Table~\ref{tab:splits}. For more details we refer the reader to~\cite{armeni_cvpr16}. Table~\ref{tab:CVPR16quantitative} tabulates the quantitative results of this baseline on the proposed dataset. Figure~\ref{fig:CVPR16qualitative} showcases sample qualitative results.

		
		\section{Conclusion}
		We presented a dataset of large-scale indoor spaces. The main property of the dataset is being comprised of mutually registered modalities including RGB images, surface normals, depths, global XYZ images, scene labels, and 2D semantics as well as raw and semantically annotated 3D meshes. The semantic annotations were performed in 3D and were consistently projected across all modalities and dimensions. We provided the 2D and 2.5D modalities in forms of both regular and $360^{\circ}$~equirectangular images. Finally, we described our collection, processing, and sampling procedures along with baseline results on 3D object detection. We hope the dataset fuels development of cross and joint modality techniques as well as unsupervised approaches leveraging the regularities in large-scale man-made spaces.

		{\small
			
		}

	\end{document}